\documentclass[runningheads,a4paper]{llncs}
\usepackage{amssymb}
\usepackage{graphicx}
\usepackage{cite}
\usepackage{amsmath,amssymb,amsfonts}
\usepackage{algorithmic}
\usepackage{adjustbox}
\usepackage{graphicx}
\usepackage{multirow}
\usepackage{textcomp}
\usepackage{lastpage}
\usepackage{ctable}
\usepackage{fancyhdr}
\usepackage{color}
\usepackage[multiple]{footmisc}

\newcommand{\rom}[1]{\uppercase\expandafter{\romannumeral #1\relax}}
\begin{document}
	
	\title{Improve the performance of transfer learning without fine-tuning using dissimilarity-based multi-view learning for breast cancer histology images}
	\titlerunning{Hongliu CAO et al.}  % abbreviated title (for running head)
	%                                     also used for the TOC unless
	%                                     \toctitle is used
	%
	\author{Hongliu CAO\inst{1,2} \and
		Simon Bernard\inst{2} \and Laurent Heutte\inst{2}\and Robert Sabourin\inst{1}}
	\author{Hongliu CAO\inst{1,2}%
		\thanks{Corresponding author, email: caohongliu@gmail.com}%
	\and
	Simon Bernard\inst{2} \and Laurent Heutte\inst{2}\and Robert Sabourin\inst{1}}
	\authorrunning{Hongliu CAO et al.} % abbreviated author list (for running head)
	%
	%%%% list of authors for the TOC (use if author list has to be modified)
	%
	\institute{Laboratoire d'Imagerie, de Vision et d'Intelligence Artificielle, \'Ecole de Technologie Sup\'erieure, \\ Universit\'e du Qu\'ebec, Montreal, Canada,\\
		\and
		Universit\'e de Rouen Normandie, LITIS (EA 4108), BP 12 - 76801 Saint-\'Etienne du Rouvray, France}

	\maketitle              % typeset the title of the contribution
	
	\begin{abstract}
		Breast cancer is one of the most common types of cancer and leading cancer-related death causes for women. In the context of ICIAR 2018 Grand Challenge on Breast Cancer Histology Images, we compare one handcrafted feature extractor and five transfer learning feature extractors based on deep learning. We find out that the deep learning networks pretrained on ImageNet have better performance than the popular handcrafted features used for breast cancer histology images. The best feature extractor achieves an average accuracy of 79.30\%. To improve the classification performance, a random forest dissimilarity based integration method is used to combine different feature groups together. When the five deep learning feature groups are combined, the average accuracy is improved to 82.90\% (best accuracy 85.00\%). When handcrafted features are combined with the five deep learning feature groups, the average accuracy is improved to 87.10\% (best accuracy 93.00\%).
		
	\end{abstract}
	\section{Introduction}
	
	The detection and treatment of cancer are still very challenging.
	The normal process of cancer detection is from certain signs and symptoms to the further investigation by medical imaging and at last confirmed by biopsy \cite{coroller2015ct,aerts2014decoding}. The diagnosis of breast cancer usually uses the biopsy tissue. The pathologists can histologically assess the microscopic structure and elements of the tissue from breast tissue biopsies \cite{araujo2017classification}.
	
	One of the most important method for tumor histological examination in pathology is Hematoxylin and eosin (H\&E) staining \cite{chan2014wonderful}. However, manual analysis is experience based, qualitative and always causes intra- or inter-observers variation even for experienced pathologists \cite{meyer2005breast}. Hence developing a more efficient, accurate, quantitative and automated system is necessary and urgent. Due to the high performance of deep learning networks, more and more studies used deep learning for the classification of breast cancer images \cite{spanhol2016breast}. However, the number of images available has always been an obstacle for the use of deep learning. Many studies divide images into patches for data augmentation, but the new problem is that there are no label information for patches. 
	
   In this paper, transfer learning without fine-tuning is proposed to solve the above problems. Six different feature extractors are compared, including five deep learning architectures and a traditional feature extractor combining  PFTAS (Parameter-Free Threshold Adjacency Statistics) and GLCM (Gray Level Co-Occurrence Matrices) features. When all features are combined, there are mainly three challenges from the machine learning point of view: (i) small sample size: size: like most
other medical applications, the number of breast cancer histology images is very
small (400 images); (ii) high dimensional feature space: as six groups of features may be combined, the size of the feature space may be up to 31855, which is over 80 times bigger than the sample size; (iii) multiple feature groups: it may be hard to improve the learning performance by exploiting the complementary information that different groups contain \cite{cao2018}. 
  To deal with these three challenges, we propose to treat breast cancer histology image classification as a multi-view learning problem. A multi-view RFSVM method proposed in our previous work\cite{cao2018} is then used as a solution. 
  
  The remainder of this paper is organized as follows: the six feature extractors are detailed in Section \rom{2}; in Section \rom{3}, the dissimilarity based multi-view learning solution is introduced; we describe in Section \rom{4} the data sets chosen in this study and provide the protocol of our experimental method; we analyze in Section \rom{5} the results of our experiments; the final conclusion and future works are drawn in Section \rom{6}.
\iffalse
We compare the performance of handcrafted features with deep learning feature extractors pretrained on ImageNet. We find out that the networks trained on ImageNet are also good feature extractors for breast cancer histology images and have better performance than handcrafted features. With random forest dissimilarity based integration method RFSVM to integrate 6 feature groups,  we are able to improve significantly the classification performance. 
, which is over 80 times bigger than the sample size
	
\fi		
	\section{Feature extractors}
	In total six different feature extractors are used in this work: handcrafted features, ResNet-18, ResNeXt, NASNet-A, ResNet-152 and VGG16. In this section, a brief introduction of each feature extractor is given. The handcrafted features include PFTAS and GLCM and have been chosen due to their good performance on breast cancer histology image classification \cite{spanhol2016dataset}. The five deep learning networks have been chosen for their performance and because they are built on different structures with different depths, and the pre-trained models are available online\footnote{https://github.com/Cadene/pretrained-models.pytorch}\footnote{https://github.com/pytorch/vision/tree/master/torchvision}.
	\subsubsection{Handcrafted features:}
	
	Two kinds of feature extractors are combined together to form the handcrafted feature group: PFTAS and GLCM. TAS (Threshold Adjacency Statistics) is a simple and fast
	morphological measure for cell phenotype
	image classification presented by Hamilton et al. in \cite{hamilton2007fast}. Similar to the work of \cite{spanhol2016dataset}, we use the Parameter-Free Threshold
	Adjacency Statistics (PFTAS) from the python library Mahotas\cite{coelho:mahotas} to build a 162-dimensional PFTAS-feature vector. GLCM features are widely used to describe the texture of tumor in cancer applications. Same as PFTAS, the library Mahotas is used to calculate the GLCM features leading to a 175-dimensional GLCM-feature vector.
	
	\subsubsection{ResNet-18 and ResNet-152:}
	ResNet is one of the deepest deep learning architectures proposed by Microsoft researchers. The deep residual nets based methods have won the first places on the tasks of ImageNet detection, ImageNet localization, COCO detection, and COCO segmentation as well as the first place on the ILSVRC 2015 classification task\cite{he2016deep}. We use two ResNet in this work: ResNet-18 and ResNet-152. Both networks take as input a \{3, 224, 224\} RGB image and are pretrained on ImageNet with 1000 classes\footnote{https://download.pytorch.org/models/resnet18-5c106cde.pth}\footnote{http://data.lip6.fr/cadene/pretrainedmodels/fbresnet152-2e20f6b4.pth}. Features are extracted from the average pool layer (i.e. before the last classification layer), which results in 512 features for ResNet-18 and 2048 features for ResNet-152.
	
	\subsubsection{ResNeXt:}
	ResNeXt is one of the state-of-the-art techniques for object recognition. It builds upon the concepts of repeating layers while exploiting the split-transform-merge strategy to bring about a new and improved architecture \cite{xie2017aggregated}. The input space of ResNeXt is a \{3, 224, 224\} RGB image and we use the network pretrained on ImageNet with 1000 classes\footnote{http://data.lip6.fr/cadene/pretrainedmodels/resnext101-64x4d-e77a0586.pth}. 2048 features are extracted from the average pool layer (i.e. before the last classification layer).
	
	\subsubsection{NASNet-A:}
	In the work of \cite{zoph2017learning}, the authors proposed to search for an architectural building block on a small dataset and then transfer the block to a larger dataset to reduce the computation cost and improve the efficiency. They used NAS (Neural Architecture Search) framework from \cite{zoph2016neural} as the main search method for their NASNets. The three networks constructed from the best three searches are named NASNet-A, NASNet-B and NASNet-C respectively. In this work, a NASNet-A pretrained on ImageNet is used.  The input space of NASNet-A is a \{3, 331, 331\} RGB image and we use the network pretrained on ImageNet with 1001 classes (ImageNet+background)\footnote{https://data.lip6.fr/cadene/pretrainedmodels/nasnetalarge-a1897284.pth}. 4032 features are extracted from the last layer before the classification layer.

	\subsubsection{VGG16:}
	The VGG Network was introduced by the researchers at Visual Graphics Group at Oxford \cite{simonyan2014very}. This network is specially characterized by its pyramidal shape.  VGG16 takes as input a \{3, 224, 224\} RGB image and we use the network pretrained on ImageNet with 1000 classes\footnote{https://download.pytorch.org/models/vgg16-397923af.pth}.  Features are extracted from the last max pooling layer, which results in 512x7x7 features.

\iffalse
    We did not choose the fully connected layer as others did because the information after the fully connected layer may be too specific to ImageNet-like problems. 
    \fi
	
	\section{ DISSIMILARITY-BASED LEARNING}
	In our previous work \cite{cao2018}, we proposed to use RFSVM to integrate information from different views together (each feature group is a view in multi-view learning framework). We have shown that RFSVM offers a good performance on Radiomics data. The RFSVM method can deal well with high dimensional low sample size multi-view data because: (i) RFSVM uses random forest dissimilarity measure to transfer each view of the data to a dissimilarity matrix so that the data dimension is reduced without feature selection, and at the same time the data in each view become directly comparable; (ii) RFSVM can take advantage of the complementary information contained in each view by combining the dissimilarity matrices together. We now recall the RFSVM method.
	
	\textbf{Random forest}: Given a training set $\mathbf{T}$, a Random Forest classifier $\mathbf{H}$ is a classifier made up of $M$ trees denoted as in Equation \eqref{e2}:
	\begin{equation}\label{e2}
	\mathbf{H}(\mathbf{X}) = \{h_k(\mathbf{X}),k=1,\dots,M\}
	\end{equation}
	where $h_k(\mathbf{X})$ is a random tree grown using the Bagging and the Random Feature Selection techniques as in \cite{biau2016random}. For predicting the class of a given query point $\mathbf{X}$ with such a tree, $\mathbf{X}$ goes down the tree structure, from its root till its terminal node. The prediction is given by the terminal node (or leaf node) in which $\mathbf{X}$ has landed. We refer the reader to \cite{biau2016random} for more information about this process. Hence if two query points land in the same terminal node, they are likely to belong to the same class and they are also likely to share similarities in their feature vectors, since they have followed the same descending path. 
	
	\textbf{Random Forest Dissimilarity (RFD)}: the RFD measure is inferred from a RF classifier $\mathbf{H}$, learned from $\mathbf{T}$. Let us firstly define a dissimilarity measure inferred by a decision tree $d^{(k)}$: let $L_k$ denote the set of leaves of the $k$th tree, and let $l_k(\mathbf{X})$ denote a function from $\mathbb{X}$ to $L_k$ that returns the leaf node of the $k$th tree where a given instance $\mathbf{X}$ lands when one wants to predict its class. The dissimilarity measure $d^{(k)}$, inferred by the $k$th tree in the forest is defined as in Equation \eqref{sk}: if two training instances $\mathbf{X}_i$ and $\mathbf{X}_j$ land in the same leaf of the $k$th tree, then the dissimilarity between both instances is set to 0, else set to 1. 
	\begin{equation}\label{sk}
	d^{(k)}(\mathbf{X}_i, \mathbf{X}_j)=
	\begin{cases}
	0, & \text{if}\ l_k(\mathbf{X}_i) = l_k(\mathbf{X}_j)\\
	1, & \text{otherwise}
	\end{cases}
	\end{equation}
	
	The RFD measure $d^{(\mathbf{H})}$ consists in calculating the $d^{(k)}$ value for each tree in the forest, and to average the resulting dissimilarity values over the $M$ trees, as in Equation \eqref{simil}:
	\begin{equation}\label{simil}
	d^{(\mathbf{H})}(\mathbf{X}_i, \mathbf{X}_j) = \frac{1}{M}\sum_{k=1}^{M} d^{(k)}(\mathbf{X}_i, \mathbf{X}_j)
	\end{equation}
	
	\textbf{Multi-view learning dissimilarities}: For multi-view learning tasks, the training set $\mathbf{T}$ is composed of $K$ views: $\mathbf{T}^{(k)} = \{(\mathbf{X}_1^{(k)}, y_1),\dots,(\mathbf{X}_N^{(k)}, y_N)\}$, k=1..K. Firstly, for each view $\mathbf{T}^{(k)}$, the RFD matrix is computed and noted as $\{\mathbf{D}_{\mathbf{H}}^k, k =1..K \}$. In multi-view learning, the joint dissimilarity matrix can typically be computed by averaging over the $K$ matrices as in Equation \eqref{av}: 
	
	\begin{equation}\label{av}
	\mathbf{D}_{\mathbf{H}} = \frac{1}{K}\sum_{i=1}^{K}\mathbf{D}_{\mathbf{H}}^i
	\end{equation}
	
	\textbf{Multi Random Forest kernel SVM (RFSVM)}: 
	From the joint RFD matrix $\mathbf{D}_{\mathbf{H}}$ of Equation \eqref{av}, one can calculate the joint similarity matrix $\mathbf{S}_{\mathbf{H}}$ as in Equation \eqref{avs}:
	\begin{equation}\label{avs}
	\mathbf{S}_{\mathbf{H}} = \mathbf{1}- \mathbf{D}_{\mathbf{H}}
	\end{equation}
	where $\mathbf{1}$ is a matrix of ones. SVM is one of the most successful classifier. Apart from the most used gaussian kernel, a lot of custom kernels can also be used: we use the joint similarity matrix $\mathbf{S}_{\mathbf{H}}$ inferred from the RF classifier $\mathbf{H}$ as a kernel in a SVM classifier.

	\section{EXPERIMENTS}
	
	The dataset used in this work is from ICIAR 2018 Grande Challenge on BreAst Cancer Histology images\footnote{https://iciar2018-challenge.grand-challenge.org/dataset/}. It is composed of Hematoxylin and eosin stained breast histology microscopy images. Microscopy images are labeled as normal, benign, in situ carcinoma or invasive carcinoma according to the predominant cancer type in each image. It is a balanced dataset with in total 400 images.

The protocol of the experiments is as follows:
\begin{itemize}
\item First, the 6 feature extractors described in Section 2 are used to extract features from histology image data. As there is no patch label provided, to simplify the feature extracting process, all images are rescaled to the network input size.
\item Second, for each group of features, a random forest with 500 trees is built. The performance of each feature group is measured by the classification accuracy of the random forest. The random forest dissimilarity matrix is calculated for each group too.
\item Finally, the RFSVM method described in Section 3 is used to combine all the groups together. Two RFSVMs are used: \textit{RFSVM (DL only)}  combines the five deep learning based feature groups; \textit{RFSVM-All} combines all the six feature groups. For RFSVM, the search range of parameter $C$ for SVM is \{0.01, 0.1, 1, 10, 100, 1000\}.
\end{itemize}

	Note that in \cite{bill2014comparative}, the authors found that when dealing with high dimensional low sample size data, stratification of the sampling is central for obtaining minimal misclassification. In this work, the stratified random splitting procedure is repeated 10 times, with 75\% as training data and 25\% as testing data. In order to compare the methods, the mean and standard deviations of accuracy were evaluated over the 10 runs. However, for the contest, only one model can be submitted. Hence the best performance among the 10 runs is also presented and chosen as the model for the contest.

	\section{Results}
	The results of the experiments are shown in Table \ref{tb1}. We can tell that the best feature extractor is ResNet-152 with an average accuracy of 79.30\% and best accuracy of 83.00\%. Followed by ResNeXt with an average accuracy of 78.60\% and the best accuracy of 81.00\%. Surprisingly, 
	the worst feature extractor is handcrafted features with PFTAS and GLCM with an average accuracy of 67.00\%. In the work of \cite{spanhol2016dataset}, PFTAS and GLCM are the best features for breast cancer histology image classification.
	By comparing the performance of the six feature extractors, we can see that even though the deep learning networks are pretrained on ImageNet dataset, which is very different from histology images, they still have a better performance as a feature extractor for breast cancer data than the best handcrafted feature extractor used in the field of breast cancer histology image classification.
	
	\begin{table}[]
		\centering
		\caption{The image wise classification results with 75\% training data and 25\% test data. \textit{Average} is the average accuracy over 10 runs, \textit{Best} is the best accuracy among the 10 runs.}
		\label{tb1}
		\begin{tabular}{p{3.5cm} p{3.5cm}l}
			\hline\noalign{\smallskip}
			& Average& Best    \\
			\noalign{\smallskip}\hline\noalign{\smallskip}
			Handcrafted &$67.00\%\pm5.46$ &$ 76.0\%$\\
			\noalign{\smallskip}\hline\noalign{\smallskip}
			ResNet-18 &$75.10\%\pm5.46$ &$ 78.0\%$\\
			\noalign{\smallskip}\hline\noalign{\smallskip}
			ResNeXt&$78.60\%\pm1.74$&$81.0\%$\\
			\noalign{\smallskip}\hline\noalign{\smallskip}
			NASNet-A &$74.70\%\pm2.33$&$78.0\%$\\
			\noalign{\smallskip}\hline\noalign{\smallskip}
			ResNet-152 &$79.30\%\pm3.20$&$ 83.0\%$\\
			\noalign{\smallskip}\hline\noalign{\smallskip}
			VGG16&$68.00\%\pm5.04$&$ 78.0\%$\\
			\specialrule{.2em}{.1em}{.1em} 
			RFSVM(DL only)&$82.90\%\pm1.37$
			&$ 85.0\%$\\
			\noalign{\smallskip}\hline\noalign{\smallskip}
			RFSVM-All&	$\textbf{87.10}\%\pm2.17$& $ \textbf{93.0}\%$ \\
			\noalign{\smallskip}\hline
			% 	LateRF		$82.20\%\pm3.12$      86
			% LateRFDIS     $81.10\%\pm3.01$	 85
			% RFDIS         $81.80\%\pm2.04$	85
			
		\end{tabular}
	\end{table}

	\begin{table}[]
		\centering
		\caption{The confusion matrix, sensitivity and specificity of our best model.}
		\label{tb2}
		\begin{tabular}{p{2.cm} p{1.5cm} p{1.5cm} p{1.5cm} p{1.5cm}}
			\hline\noalign{\smallskip}
			& Benign& InSitu &Invasive & Normal    \\
			\noalign{\smallskip}\hline\noalign{\smallskip}
			Benign &23 & 1 & 0&  1\\
			\noalign{\smallskip}\hline\noalign{\smallskip}
			InSitu &0&23&  0  &2\\
			\noalign{\smallskip}\hline\noalign{\smallskip}
			Invasive&1&  0& 24 & 0\\
			\noalign{\smallskip}\hline
			Normal&2 & 0&  0 &23\\
            \specialrule{.2em}{.1em}{.1em} 
			Sensitivity&92\% & 92\%& 96\% &92\%\\
			\noalign{\smallskip}\hline
			Specificity&85\% &96\%&  100\% &85\%\\
			\noalign{\smallskip}\hline
			% 	LateRF		$82.20\%\pm3.12$      86
			% LateRFDIS     $81.10\%\pm3.01$	 85
			% RFDIS         $81.80\%\pm2.04$	85
			
		\end{tabular}
	\end{table}	 	
	With \textit{RFSVM (DL only)} integrating all the five deep learning based feature groups together, the average accuracy is improved to 82.90\% and the best performance is improved to 85.00\%. However, when all feature groups are combined with \textit{RFSVM-All}, the average accuracy is improved to 87.10\% and the best performance is improved to 93.00\%. It shows that even though the handcrafted features do not have a very good performance individually, they can still provide useful complementary information for breast cancer classification when combined with deep learning based feature groups.
	
	The confusion matrix, sensitivity and specificity of our best model are shown in Table \ref{tb1}. From the results we can see that our model has very high sensitivity on all four classes, and very high specificity too for two classes, i.e. \textit{InSitu} and \textit{Invasive}.
	
	Note that the state of the art performance on this dataset is considered to be from \cite{araujo2017classification}. In this work, the authors used CNN patch-wise training on a previous version of the dataset with 249 images for training and 20 images for testing (7.4\% of the whole dataset as test data). They obtained as best performance an accuracy of 85.00\%. In our work, 300 images are used for training and 100 images are used as test data. Hence, even if the results are not directly comparable with \cite{araujo2017classification}, the accuracy of our best model is 8\% higher than the accuracy reported in \cite{araujo2017classification} while using 25\% of the whole dataset as test data, which is much more than 7.4\% in \cite{araujo2017classification}.

	\section{Conclusion}
	In this work, we firstly compared the popular handcrafted features used in breast cancer histology image classification with five deep learning based feature extractors pretrained on ImageNet. Not surprisingly, the experimental results show that the deep learning based features are better than the handcrafted. To improve the performance of transfer learning, we tackled the problem of breast cancer histology image classification as an HDLSS multi-view learning task and applied an RFSVM method previously proposed for the classification of Radiomics data. The results obtained with \textit{RFSVM (DL only)} show that the performance of transfer learning can be improved by combining multiple feature extractors together. The results obtained with \textit{RFSVM-All} show that even though deep learning based features have better performance than handcrafted features for breast cancer histology image classification, the accuracy can be improved significantly when they are combined together and surpass the state of the art performance on the dataset used. 
	
	\section*{Acknowledgment}
This work is part of the DAISI project, co-financed by the European Union with the European Regional Development Fund (ERDF) and by the Normandy Region.

	\bibliographystyle{ieeetr}
	\bibliography{sample}
\end{document}